\documentclass[letterpaper]{article} 
\usepackage{aaai25}  
\usepackage{times}  
\usepackage{helvet}  
\usepackage{courier}  
\usepackage[hyphens]{url}  
\usepackage{graphicx} 
\usepackage{natbib}  
\usepackage{caption} 
\usepackage{enumitem}
\usepackage{romannum}
\usepackage{booktabs}
\usepackage{amsmath}
\usepackage{cuted}
\usepackage{capt-of}

\usepackage{array}
\newcolumntype{P}[1]{>{\centering\arraybackslash}p{#1}}
\usepackage{amssymb}
\usepackage{xcolor,colortbl}
\definecolor{Gray}{gray}{0.9}
\usepackage{cite}
\usepackage{algorithm}
\usepackage{algpseudocode}

\usepackage{newfloat}
\usepackage{listings}
\DeclareCaptionStyle{ruled}{labelfont=normalfont,labelsep=colon,strut=off} 

\floatstyle{ruled}
\newfloat{listing}{tb}{lst}{}
\floatname{listing}{Listing}
\title{Elevating Flow-Guided Video Inpainting with Reference Generation}

\author{
Suhwan Cho$^{1,}$\thanks{This work was done during an internship at Adobe Research.}\quad Seoung Wug Oh$^2$\quad Sangyoun Lee$^1$\quad Joon-Young Lee$^2$
}

\affiliations{
\vspace{2mm}
$^1~$Yonsei University, Seoul, Korea\\
\vspace{0.5mm}
$^2~$Adobe Research, San Jose, California, USA\\
\vspace{0.5mm}
\{chosuhwan, syleee\}@yonsei.ac.kr\quad \{seoh, jolee\}@adobe.com
}

\begin{document}
\maketitle

\begin{abstract}
Video inpainting (VI) is a challenging task that requires effective propagation of observable content across frames while simultaneously generating new content not present in the original video. In this study, we propose a robust and practical VI framework that leverages a large generative model for reference generation in combination with an advanced pixel propagation algorithm. Powered by a strong generative model, our method not only significantly enhances frame-level quality for object removal but also synthesizes new content in the missing areas based on user-provided text prompts. For pixel propagation, we introduce a one-shot pixel pulling method that effectively avoids error accumulation from repeated sampling while maintaining sub-pixel precision. To evaluate various VI methods in realistic scenarios, we also propose a high-quality VI benchmark, HQVI, comprising carefully generated videos using alpha matte composition. On public benchmarks and the HQVI dataset, our method demonstrates significantly higher visual quality and metric scores compared to existing solutions. Furthermore, it can process high-resolution videos exceeding 2K resolution with ease, underscoring its superiority for real-world applications. Code and models are available at \url{https://github.com/suhwan-cho/RGVI}.
\end{abstract}

\begin{figure*}[t]
\centering
\includegraphics[width=1\linewidth]{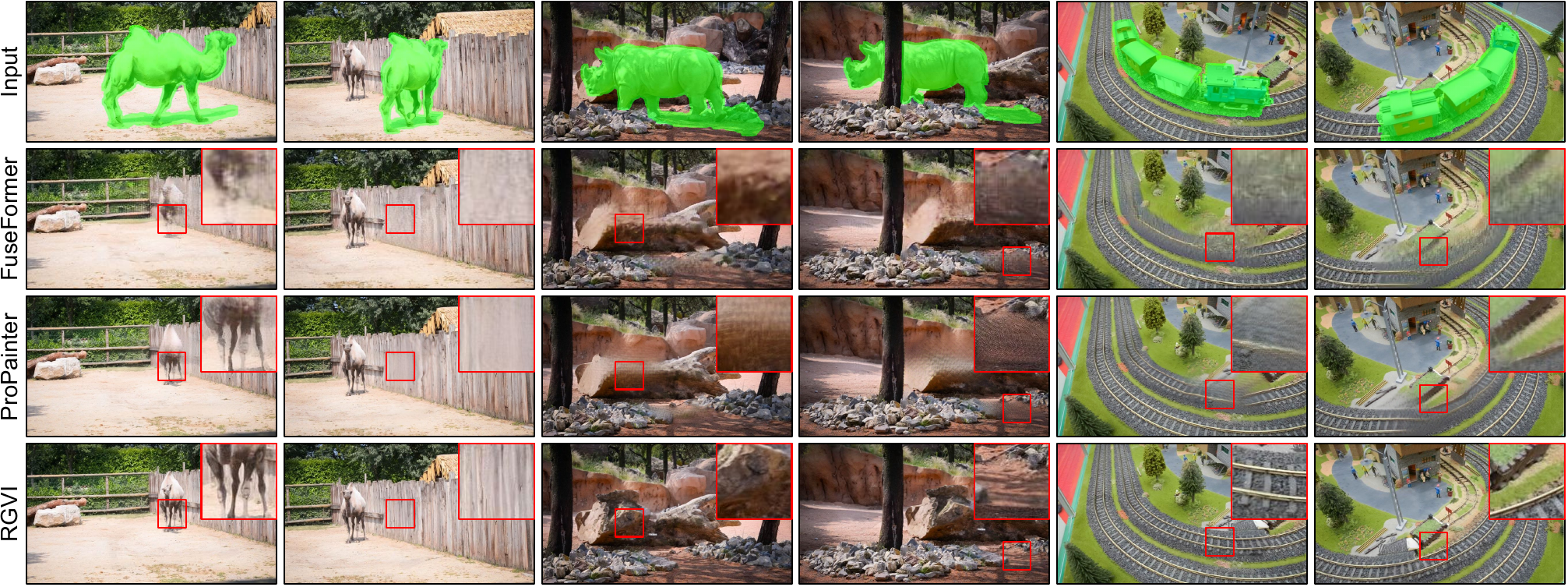}
\caption{Qualitative comparison between RGVI and state-of-the-art methods.}
\label{figure1}
\end{figure*}

\section{Introduction}
With the recent growth of video content creation and consumption, there is a high demand for technology that can edit original video content. Video inpainting (VI) is one of the essential video editing steps that allows content creators to remove certain areas or objects from existing videos and fill them with plausible content. While image inpainting has recently made breakthroughs based on diffusion models, VI technology has not kept pace. The challenge is that even with strong generation capabilities, maintaining original video content and temporal consistency remains difficult. To design a powerful VI system, the algorithm needs to effectively propagate observable content across frames while simultaneously generating new, non-observable content that does not appear in the video.

Most existing methods approach propagation and generation in a coupled manner through end-to-end training. Earlier works train a generative model based on 3D convolutions~\cite{CombCN, FVI, LGTSM} using adversarial loss. Although these methods are plausible for hallucination, they often fail to maintain temporal consistency due to limited temporal window size. Some attempts address this limitation with temporal relation reasoning through attention mechanisms~\cite{OPN, STTN, FuseFormer} or homography transformation~\cite{CPNet, CANet, VIST}. However, these methods fail to generate plausible content when there is no reference available in the video. In a coupled approach, there is always ambiguity between propagation and generation for the best results, making it difficult to balance the two.

Conversely, flow-based methods~\cite{DFVI, ECFVI, VINet} achieve great success in producing high-quality VI results by adopting a decoupled framework. They compute optical flows to propagate content between frames and utilize a separate image inpainting model to synthesize non-observable content. However, the difficulty of pixel propagation implementation becomes a bottleneck for flow-based approaches, affecting both quality and efficiency. For instance, FGVC~\cite{FGVC} uses a per-pixel flow tracing algorithm that loses sub-pixel accuracy. E$^2$FGVI~\cite{E2FGVI} and ProPainter~\cite{ProPainter} use a recurrent pixel warping algorithm that preserves sub-pixel accuracy but introduces re-sampling artifacts due to repeated color sampling.

In this study, we propose a practical and powerful VI system named reference-guided video inpainting (RGVI), which overcomes the limitations of existing flow-based approaches. We introduce a new pixel propagation algorithm by combining flow tracing and grid warping to avoid re-sampling artifacts while maintaining sub-pixel accuracy. Specifically, we warp optical flows instead of color values and pull the color value from the matching pixel in a one-shot manner. Additionally, we propose a propagation verification method to detect areas where propagation is unreliable. We also introduce a simple and effective trick to prevent common errors in flow-based approaches by providing the mask of the occluding object, thus preventing color bleeding artifacts from inaccurate optical flows.

We argue that the potential of our decoupled framework is fully realized when used in conjunction with a strong generative model for reference generation. We adopt Stable Diffusion based on the latent diffusion model~\cite{LDM}, which not only greatly improves inpainting quality but also offers texture replacement based on text guidance. With this combination, our method can produce high-quality results with high controllability. Thanks to the strong generative capability, it can handle 2K resolution videos with ease, even when large missing area completion is required, significantly increasing its usability.

To quantitatively evaluate our method on realistic videos, we propose a high-quality VI benchmark dataset named HQVI. Instead of randomly corrupting videos with random objects or free-form masks~\cite{STTN, FuseFormer, E2FGVI}, we carefully design each video sequence by blending foreground objects with the background video using alpha matte composition. This approach simulates realistic video editing scenarios while providing ground truth for object removal. To consider real-world use cases, we include generation-required scenarios where large missing area completion is necessary. The dataset also contains a new VI setting where the target object to be removed is interacting with distracting objects. To facilitate future research, we provide a comprehensive evaluation of existing VI approaches on the HQVI dataset.

On public benchmarks, RGVI showcases notably superior visual quality and temporal consistency, as depicted in Figure~\ref{figure1}. Quantitative evaluations further substantiate its efficacy. Our main contributions can be summarized as follows:
\begin{itemize}[leftmargin=0.2in]
\item We introduce a robust VI framework that integrates a large generative model by decoupling the inherent ambiguity between content propagation and generation.

\item We present a novel pixel propagation algorithm that integrates flow tracing and grid warping, which ensures sub-pixel accuracy while mitigating re-sampling artifacts.

\item We introduce the HQVI benchmark dataset to assess VI performance in realistic scenarios, which includes a comprehensive evaluation of our proposed method alongside existing solutions.
\end{itemize}

\begin{figure*}[t]
\centering
\includegraphics[width=1\linewidth]{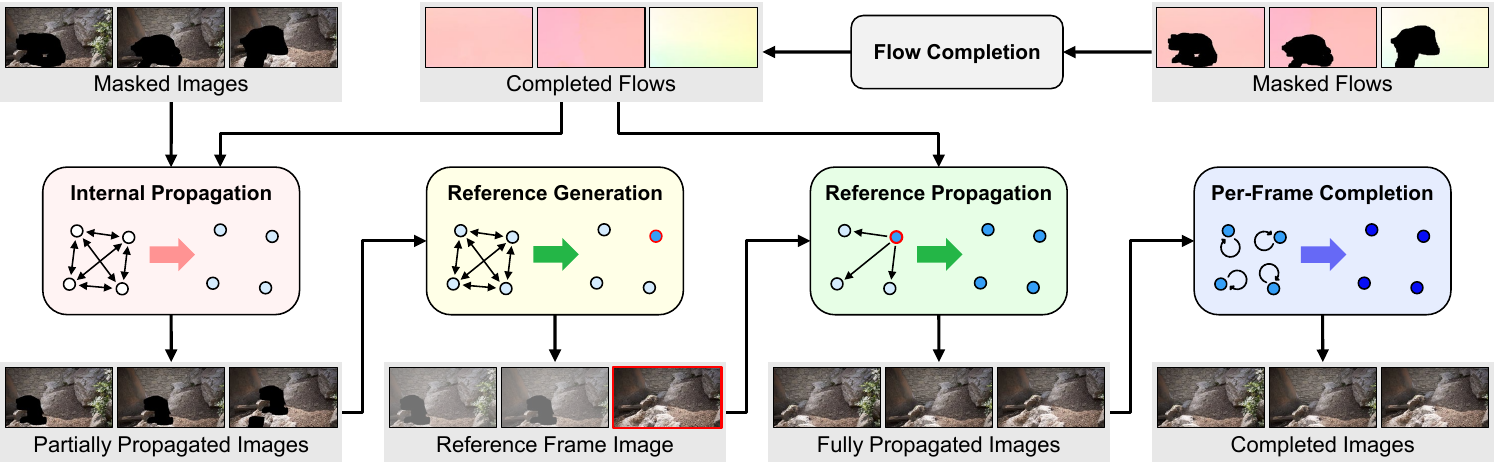}
\caption{Overall pipeline of RGVI.}
\label{figure2}
\end{figure*}

\section{Related Work}
\noindent\textbf{End-to-end approaches.} Some methods employ an end-to-end framework where propagation and generation are integrated into a single generative model trained using adversarial loss. CombCN~\cite{CombCN}, FVI~\cite{FVI}, and LGTSM~\cite{LGTSM} extend image inpainting techniques to the video domain, incorporating temporal considerations through 3D convolutions. OPN~\cite{OPN}, STTN~\cite{STTN}, FuseFormer~\cite{FuseFormer}, DLFormer~\cite{DLFormer}, and FGT~\cite{FGT} utilize attention mechanisms~\cite{transformer, ViT} for implicit relation reasoning among embedded features, facilitating spatio-temporal context propagation.

However, achieving effective propagation and generation simultaneously remains challenging for these methods, often resulting in a balance issue between the two tasks. Moreover, they may struggle to fully leverage the capabilities of foundational generative models due to the complexity of handling both tasks concurrently. Additionally, approaches like InterVI~\cite{InterVI} and IIVI~\cite{IIVI} adopt internal learning strategies to adapt the network to specific videos at test time, yet they face similar challenges in achieving optimal balance and integration between propagation and generation phases.

\vspace{1mm}
\noindent\textbf{Flow-based approaches.} Several existing methods exploit the inherent visual similarity across video frames by adopting flow-based frameworks. DFVI~\cite{DFVI} and FGVC~\cite{FGVC} utilize optical flows to directly allocate pixel values, propagating known pixels into missing regions. In contrast, methods like VINet~\cite{VINet}, TSAM~\cite{TSAM}, and E$^2$FGVI~\cite{E2FGVI} propagate features rather than pixels between adjacent frames, followed by feature decoding. ProPainter~\cite{ProPainter} combines optical flow warping with the propagation of both pixels and features, effectively integrating pixel-level details with feature-level semantic information.

Flow-based methods typically decouple propagation and generation modules, often yielding clearer results compared to end-to-end methods. However, they are not without limitations. For instance, FGVC's per-pixel forward flow tracing lacks sub-pixel accuracy, leading to spatial misalignment. Conversely, E$^2$FGVI and ProPainter achieve sub-pixel accuracy with a recurrent pixel warping approach but suffer from repetitive re-sampling, which can diminish detail retention. In contrast to these methods, we propose a one-shot pixel pulling algorithm that mitigates these limitations. Additionally, our approach includes a verification method to detect and correct potential errors based on propagation reliability.

\section{Approach}
\subsection{System Overview}
Given input images and binary masks, the objective of VI is to erase the masked regions and fill them with plausible content. As visualized in Figure~\ref{figure2}, RGVI consists of four key components: 1) internal pixel propagation to fill the missing areas using known pixels from the video; 2) reference generation to produce high-quality reference content using a large generative model; 3) reference propagation to distribute the generated pixels across all frames; and 4) per-frame completion to address any remaining missing areas.

\subsection{Internal Pixel Propagation}
The missing areas in one frame can be partially completed by propagating known pixels from other frames. In the internal pixel propagation stage, known pixels are propagated across all frames based on predicted optical flows.

\vspace{1mm}
\noindent\textbf{Flow preparation.} To propagate the known pixels across frames, we first calculate the optical flows of each frame using RAFT~\cite{RAFT}. Since the predicted flows contain information about the object to be removed, we erase the flows in the masked area and complete them plausibly. For flow completion, we adopt the recurrent protocol from ProPainter~\cite{ProPainter}. This process results in completed flows $f_{i \rightarrow j}$, where $i$ and $j$ are adjacent frames.

\vspace{1mm}
\noindent\textbf{One-shot pulling via flow tracing.} We adopt a recurrent grid warping strategy for pixel propagation, allowing us to trace the optical flow with sub-pixel accuracy. The optical flows are sequentially chained as follows:
\begin{align}
&f_{i \rightarrow j} =
\begin{cases}
f_{i \rightarrow j - 1} + w(f_{j-1 \rightarrow j}, f_{i \rightarrow j - 1}) &i < j\\
f_{i \rightarrow j + 1} + w(f_{j+1 \rightarrow j}, f_{i \rightarrow j + 1}) &i > j~,
\end{cases}
\end{align}
where $w(A,B)$ is a grid warping operation that warps $A$ using flow $B$, and $i$ and $j$ are two arbitrary frames in a video. After this step, we establish a global correspondence map that aligns any source frame to the target frame. This global correspondence map is then used to pull pixel colors from the source frames to fill the target frames.

Note that while we use the recurrent grid warping operation similar to existing flow-based methods~\cite{DFVI, E2FGVI, ProPainter}, our goal is different. Our method warps flow maps, whereas other methods warp pixel color values directly. Although recurrent pixel warping works fine for short-term propagation, it naturally accumulates propagation errors in the long term, leading to re-sampling artifacts, as reported in FGVC~\cite{FGVC}. We also observe that directly propagating color values results in blurry textures. In contrast, optical flow is inherently smoother, making it more robust to re-sampling artifacts. A comparison between recurrent pixel warping and our one-shot pixel pulling is provided in Section~\ref{analysis}.

In addition to reducing re-sampling artifacts through flow tracing, our propagation method is more precise than the pixel-wise flow tracing used in FGVC, as we maintain sub-pixel accuracy. Furthermore, it is a more efficient and GPU-friendly operation, enabling our method to handle high-resolution videos within a reasonable computing time.

\vspace{1mm}
\noindent\textbf{Bi-directional collection with verification.} In this step, we pull known pixels from the source frames to fill the missing area in the target frame. We employ two sequential passes, starting from the target frame in both forward and backward directions, to gain a global perception of the video. Our approach prioritizes pixel color values from the nearest frames. Thus, in each pass, we greedily collect pixel color values for the missing area in the target frame. Importantly, during the loop through different source frames, we ensure that pixel colors are pulled in a one-shot manner, eliminating the need for repeated sampling processes.

Once we locate the corresponding pixel in both forward and backward directions, we implement a verification protocol. Our method employs a straightforward verification approach based on color value differences. Specifically, we compute the L1 distance between the normalized 3-channel color values, ranging from 0 to 1. If the difference between the pulled colors from both directions falls below a threshold, we assign the average value to the target pixel location. Conversely, if there is disagreement (i.e., the difference exceeds the threshold), we identify these target pixels as unreliable and invalidate the propagation in subsequent steps. Note that the threshold value is empirically set to 1, with minimal observed variation across different values.

\begin{algorithm}[t!]
\small
\caption{Bi-Directional Pixel Collection}
\begin{algorithmic}[1]
\State \textbf{Input:} $X$, $M$, $f$
\State \textbf{Output:} $\hat{X}$, $\hat{M}$, $V$
\State Initialize $\hat{X}$ and $\hat{M}$ with $X$ and $M$
\For{target frame $i$}
\For{source frame $j$ larger than $i$}
\State Attach $w(X_j, f_{i \rightarrow j})$ on $\hat{X}_i^f$
\State Update $\hat{M}_i^f$
\EndFor
\For{source frame $j$ smaller than $i$}
\State Attach $w(X_j, f_{i \rightarrow j})$ on $\hat{X}_i^b$
\State Update $\hat{M}_i^b$
\EndFor
\State Compare $\hat{X}_i^f$ and $\hat{X}_i^b$
\State Calculate $\hat{X}_i, \hat{M}_i$, and $V_i$
\EndFor
\end{algorithmic}
\label{algorithm1}
\end{algorithm}

For clarity, we also provide pseudo code in Algorithm~\ref{algorithm1}. The proposed bi-directional pixel collection based on one-shot pixel pulling protocol takes masked images $X$, given masks $M$, and completed flows $f$ as input. In each target frame, known pixels from the source frames are propagated using a one-shot warping approach until the target frame is fully completed or there are no more source frames available. After completing all propagation steps, we obtain updated images $\hat{X}$, updated masks $\hat{M}$, and an invalid propagation area indicator $V \in \{0, 1\}$.

\subsection{Reference Generation and Propagation}
While we complete most of the missing areas through internal pixel propagation, some portions may remain unresolved due to insufficient intra-video information. To address these gaps, we employ a large generative model to generate reference contents, which are then propagated to other frames.

\vspace{1mm}
\noindent\textbf{Key frame selection.} To mitigate content conflicts across frames, we adopt a strategy where new contents are generated solely for a single key frame instead of independently generating them for each frame. Selecting the key frame is crucial as its generated contents will influence the entire video. We determine the key frame based on its impact after reference propagation, aiming to maximize the utilization of high-quality generated contents. Specifically, the influence of a frame $i$ is quantified by its connection count $C_i$, which measures the number of connections to unknown pixels in other frames. Formally, the connection count $C_i$ for frame $i$ is computed as follows:
\begin{align}
&C_i = \sum_{j=1}^{L}\Bigl\{\sum_p{\bigl(w(\hat{M}_j,f_{i \rightarrow j}) \odot \hat{M}_i\bigl)}\Bigl\}~,
\end{align}
where $p$ denotes the pixel index and $L$ denotes the total length of the video. Finally, based on the connection count of each frame, the key frame $k$ is defined as the frame that maximizes the connection count:
\begin{align}
&k = \underset{i}{\arg\max} C_i~.
\end{align}

While a single key frame can typically complete most of the missing areas, there are rare cases where a single reference frame is insufficient. In such instances, a straightforward solution involves sequentially performing reference generation and propagation with multiple key frames.

\begin{figure}[t!]
\centering
\includegraphics[width=1.0\linewidth]{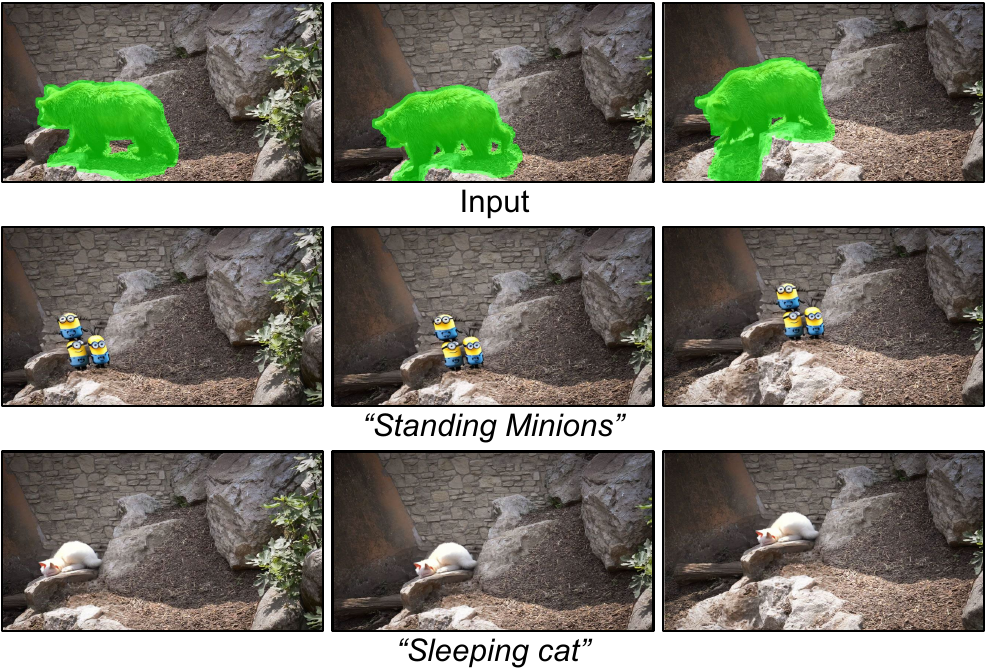}
\caption{RGVI outputs from the generation mode.}
\label{figure3}
\end{figure}

\vspace{1mm}
\noindent\textbf{Reference generation.} After selecting frame $k$ as the key frame, we generate high-quality reference contents using a large generative model. Specifically, we employ the Stable Diffusion based on latent diffusion model~\cite{LDM}. This allows us to control the generation process by providing additional text prompts. We utilize two modes for reference generation: removal mode and generation mode.

In the removal mode, which is our default setting, the goal is to produce the most plausible contents that blend seamlessly with the original images, such as the background. To avoid generating unwanted objects or patterns, we use \textit{``Empty background, high resolution"} as the text prompt for the removal mode. In contrast, in the generation mode, we use a cropped image near the missing area as the input image, thereby excluding external guidance from known pixels. This approach is intended to facilitate content generation based solely on the local context. We showcase example results of RGVI using the generation mode in Figure~\ref{figure3}.

\vspace{1mm}
\noindent\textbf{Reference propagation.} The generated pixels in the key frame $k$ are then propagated to the rest of the frames using the grid warping operation with pre-calculated flows:
\begin{align}
&\tilde{X_i} = \hat{X}_i + \hat{M}_i \odot w(\hat{X}_k, f_{i \rightarrow k})~,
\end{align}
where $\tilde{X}$ denotes the set of images after reference propagation. Similarly, we obtain a set of masks indicating areas that are still not completed, denoted as $\tilde{M}$.

\subsection{Per-Frame Completion}
After completing the previous steps, there may still be areas that remain incomplete despite efforts with known and generated pixels. Additionally, some internally propagated pixels are deemed unreliable during the propagation verification step. These unreliable pixels cannot be adequately filled using inter-frame knowledge alone, necessitating a per-frame completion strategy. Inspired by image inpainting techniques such as \cite{PartialConv, DeepFill, DeepFillv2}, we address each frame's missing areas individually. To achieve this, we employ a lightweight convolutional network structured on a simple encoder-decoder architecture. Leveraging the per-frame completion network $\Psi$, we produce a set of fully restored output images as
\begin{align}
&\Psi(\tilde{X} \odot (1 - V), \tilde{M} + V)~.
\end{align}

\vspace{1mm}
\noindent\textbf{Network training.} To train the network, we utilize images from the YouTube-VOS 2018 dataset~\cite{YTVOS2018} training set, resized to a resolution of 240$\times$432. Images are randomly selected and masked for training purposes, where the original images serve as ground truth, and the masked versions are used as inputs alongside binary masks. Our image corruption strategy includes two approaches: 1) random region free-form masking, akin to standard inpainting tasks; and 2) random object masking to simulate scenarios involving object removal. For training, we employ a straightforward combination of L1 loss and adversarial loss functions, optimized using the Adam optimizer~\cite{adam} with a fixed learning rate of 1e-4.

\subsection{Handling Occlusion with Additional Mask}
\label{mask}
Due to their structures, flow-based VI methods are often susceptible to occlusions. Ideally, optical flows should be completed consistently with the background contents. However, occluding objects' motion can significantly disrupt flow completion, leading to critical propagation errors. To mitigate this, we propose a simple yet effective technique that enhances the stability of flow-based VI methods.

In addition to providing the mask of the target object to be removed (negative mask), we also include the mask of the object occluding the target (positive mask). Before inference, we combine these masks into a temporary target mask. After inference, we overlay the original contents of the positive mask onto the output images, ensuring only the target object is removed from the video. This straightforward approach effectively addresses occlusion issues in flow-based VI methods. The impact of this technique is analyzed and discussed in Section~\ref{analysis}.

\begin{figure}[t!]
\centering
\includegraphics[width=1.0\linewidth]{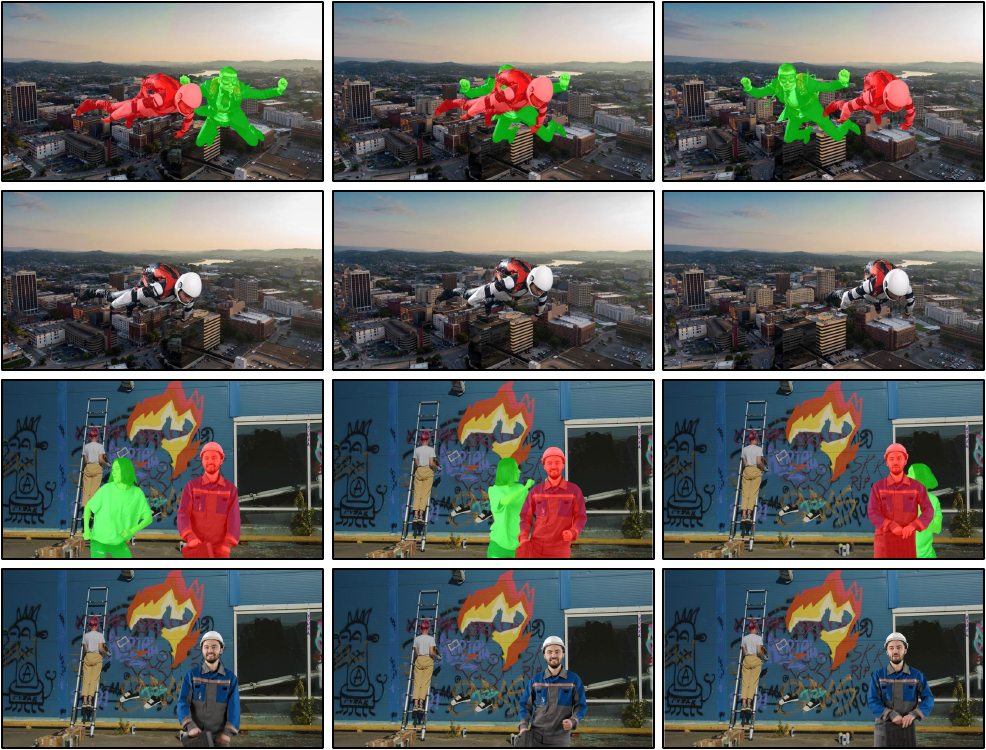}
\caption{Example videos from the HQVI dataset. Negative masks are highlighted in green, while positive masks are highlighted in red.}
\label{figure4}
\end{figure}

\section{HQVI Dataset}
We introduce HQVI, a high-quality and realistic VI benchmark dataset. Input-output pairs are synthesized using objects from VideoMatte240K~\cite{VM240K} and real-world videos from Pexels\footnote{\url{https://www.pexels.com}}, employing alpha compositing. HQVI differs from existing VI benchmark datasets in several ways: 1) it includes scenarios requiring generation with large missing areas, reflecting real-world use cases; 2) foreground-background composition employs fine alpha mattes to prevent composition artifacts, with these accurate annotations potentially serving as an additional cue for future research; 3) it incorporates target occlusion scenarios with detailed annotations for both the target object to be removed (negative mask) and the occluding object to be preserved (positive mask); 4) each video boasts a resolution of 1200$\times$2160, which is notably higher than that of other datasets.

\begin{table*}[t!]
\centering
\small
\begin{tabular}{p{2cm}P{1.6cm}P{1.8cm}P{1.1cm}P{1.1cm}P{1.1cm}P{1.1cm}P{1.1cm}P{1.1cm}}
\toprule 
Method &Publication &Resolution &PSNR$\uparrow$ &SSIM$\uparrow$ &LPIPS$\downarrow$ &VFID$\downarrow$ &Mem. &Time\\
\midrule
STTN$^\dagger$ &ECCV'20 &240$\times$432 &29.64 &0.9339 &0.0528 &0.2594 &5.2G &9s\\
FGVC &ECCV'20 &240$\times$432 &28.37 &0.9383 &0.0409 &0.2436 &1.3G &2m 56s\\
FuseFormer$^\dagger$ &ICCV'21 &240$\times$432 &29.92 &0.9365 &0.0498 &0.2727 &21.5G &17s\\
E$^2$FGVI$^\dagger$ &CVPR'22 &240$\times$432 &30.63 &0.9427 &0.0401 &0.1885 &7.2G &12s\\
ProPainter &ICCV'23 &240$\times$432 &30.62 &0.9413 &0.0388 &0.2128 &8.1G &15s\\
\rowcolor{Gray} RGVI w/o Ref. & &240$\times$432 &\textbf{31.60} &\textbf{0.9559} &0.0390 &0.1868 &8.3G &55s\\
\rowcolor{Gray} RGVI & &240$\times$432 &30.66 &0.9527 &\textbf{0.0335} &\textbf{0.1825} &8.3G &58s\\
\hline
FGVC &ECCV'20 &480$\times$864 &28.63 &0.9433 &0.0388 &0.0470 &2.4G &15m 7s\\
ProPainter &ICCV'23 &480$\times$864 &30.69 &0.9414 &0.0457 &0.0478 &23.7G &1m 12s\\
\rowcolor{Gray} RGVI w/o Ref. & &480$\times$864 &\textbf{31.19} &\textbf{0.9534} &0.0403 &0.0404 &8.3G &1m 38s\\
\rowcolor{Gray} RGVI & &480$\times$864 &30.90 &0.9513 &\textbf{0.0342} &\textbf{0.0311} &8.3G &1m 41s\\
\hline
\rowcolor{Gray} RGVI w/o Ref. & &1200$\times$2160 &29.81 &\textbf{0.9501} &0.0403 &0.0101 &17.2G &7m 56s\\
\rowcolor{Gray} RGVI & &1200$\times$2160 &\textbf{30.10} &0.9489 &\textbf{0.0357} &\textbf{0.0058} &17.2G &7m 59s\\
\hline
\end{tabular}
\caption{Quantitative comparison on the HQVI dataset, detailing their input resolutions and indicating whether a fixed input resolution is required (marked with $\dagger$). Mem. denotes maximum GPU memory usage during inference, while Time represents average inference time per sequence.}
\label{table1}
\end{table*}

\section{Experiments}
We conduct extensive experiments on the HQVI, DAVIS 2016~\cite{DAVIS2016}, and YouTube-VOS 2018~\cite{YTVOS2018} datasets to validate the effectiveness of our proposed approach. We adopt PSNR, SSIM~\cite{SSIM}, LPIPS~\cite{LPIPS}, and VFID~\cite{vfid} as evaluation metrics, utilizing AlexNet~\cite{alexnet} for feature extraction in LPIPS calculations. All experiments are conducted on a single TITAN RTX GPU. Unless otherwise specified, additional masking (Section~\ref{mask}) is not employed.

\begin{table}[t!]
\centering
\small
\begin{tabular}{lcccccc}
\toprule 
\multicolumn{1}{c}{} &\multicolumn{2}{c}{DAVI} &\multicolumn{2}{c}{YTVI}\\
\cmidrule{2-5}
Method &PSNR$\uparrow$ &SSIM$\uparrow$ &PSNR$\uparrow$ &SSIM$\uparrow$\\
\midrule
STTN &26.98 &0.8476 &29.96 &0.9102\\
FuseFormer &27.41 &0.8583 &29.18 &0.9001\\
E$^2$FGVI &28.71 &0.8885 &31.18 &0.9280\\
ProPainter &29.55 &0.9058 &\textbf{31.70} &0.9312\\
\rowcolor{Gray} RGVI w/o Ref. &\textbf{29.75} &\textbf{0.9186} &\textbf{31.70} &\textbf{0.9335}\\
\hline
\end{tabular}
\caption{Quantitative comparison on the DAVI and YTVI datasets with a 240p input resolution.}
\label{table2}
\end{table}

\begin{figure}[t]
\centering
\includegraphics[width=1.0\linewidth]{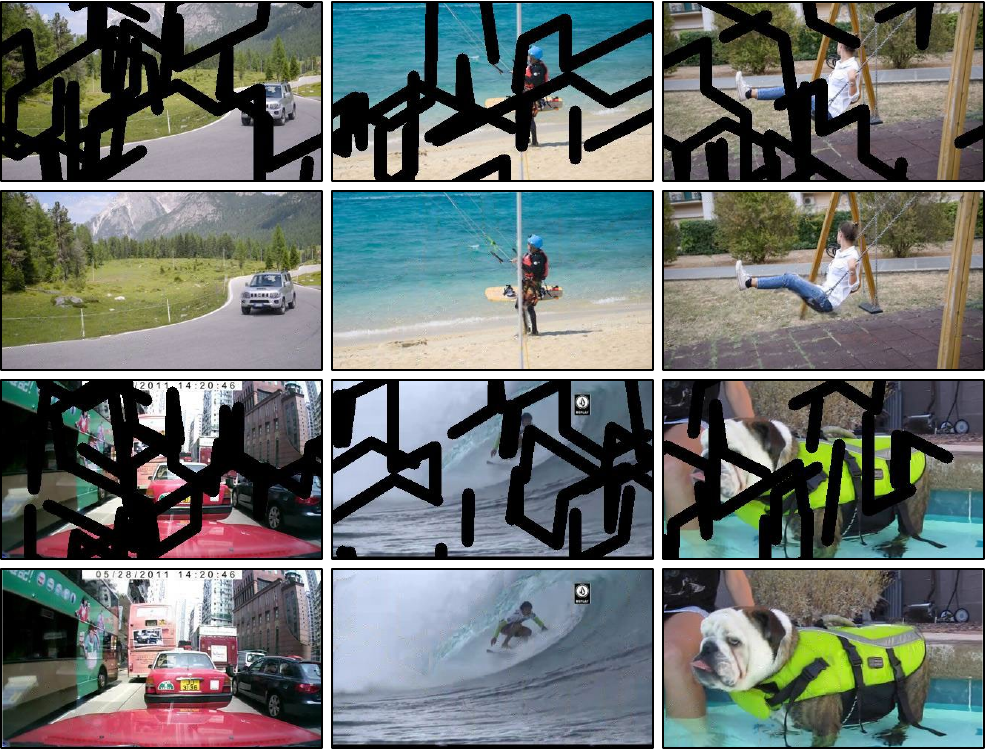}
\caption{RGVI outputs on the video restoration scenarios.}
\label{figure5}
\end{figure}

\subsection{Quantitative Results}
\label{quanti}
In Table~\ref{table1}, we quantitatively compare our proposed RGVI with state-of-the-art methods on the HQVI dataset. Our method significantly outperforms all existing methods in the video object removal setting. At 240p and 480p input resolutions, RGVI without reference achieves the highest performance on traditional metrics (PSNR and SSIM). However, on the perceptual metrics (LPIPS and VFID), RGVI with reference achieves the best scores. This discrepancy indicates that high-fidelity images often receive low scores on traditional metrics, which fail to properly evaluate generation quality, as blurry textures tend to score higher. Notably, for 2K resolution videos with large missing areas, RGVI with reference outperforms RGVI without reference even in PSNR, highlighting the necessity of large generative models. Unlike conventional VI solutions that overlook high-resolution use cases, RGVI demonstrates significant advantages in addressing such scenarios.

Regarding scalability, existing methods fail to handle 2K resolution with common GPUs, often resulting in GPU memory overflow~\cite{FGVC, ProPainter}. Some methods~\cite{STTN, FuseFormer, E2FGVI} are strictly trained on specific resolutions and do not scale to higher resolutions. Unlike these methods, RGVI demonstrates high scalability with an excellent performance-to-cost ratio, completing the entire process in a reasonable time even for 2K video input.

Following the common evaluation protocol in VI~\cite{CombCN, FVI, SSVI}, we also compare various methods under the video restoration setting in Table~\ref{table2}. We prepare two commonly adopted datasets: a combination of the DAVIS 2016 training set and validation set (50 videos) and the YouTube-VOS 2018 testing set (508 videos). The images in each dataset are corrupted using random free-form masking, with the corrupted images serving as input and the original images as ground truth. We refer to the constructed datasets as DAVI and YTVI, respectively. Since reference generation is not required in most video restoration cases, we skip the reference generation process of RGVI. The table shows that RGVI achieves satisfactory performance on both the DAVI and YTVI datasets, supporting its efficacy for video restoration scenarios as well.

\subsection{Qualitative Results}
Qualitative evaluation is crucial for assessing VI performance, as visual quality is sometimes not well reflected in quantitative metrics. In Figure~\ref{figure1}, we show video object removal examples on the DAVIS 2016 dataset, where the object masks are dilated following the prior work~\cite{dilate}. Compared to FuseFormer~\cite{FuseFormer} and ProPainter~\cite{ProPainter}, RGVI demonstrates significantly higher performance in propagating observable contents within a video and generating new content that does not exist in the video. In Figure~\ref{figure5}, we present the video restoration results on the DAVI and YTVI datasets. RGVI is also highly effective for video restoration scenarios.

\begin{table}[t]
\centering
\small
\begin{tabular}{P{0.8cm}|P{0.8cm}|cccc}
\toprule 
Int. &Ref. &PSNR$\uparrow$ &SSIM$\uparrow$ &LPIPS$\downarrow$ &VFID$\downarrow$\\
\midrule
- &- &27.64 &0.9234 &0.0962 &0.5916\\
\midrule
Rec. &- &31.43 &0.9512 &0.0595 &0.2405\\
One. &- &31.60 &0.9559 &0.0390 &0.1868\\
\midrule
Rec. &Rec. &30.17 &0.9478 &0.0558 &0.2153\\
One. &One. &30.66 &0.9527 &0.0335 &0.1825\\
\bottomrule
\end{tabular}
\caption{Quantitative analysis of pixel propagation strategies on the HQVI dataset with a 240p input resolution.}
\label{table3}
\end{table}

\begin{table}[t]
\centering
\small
\begin{tabular}{c|cccc}
\toprule 
Add. Mask &PSNR$\uparrow$ &SSIM$\uparrow$ &LPIPS$\downarrow$ &VFID$\downarrow$\\
\midrule
&35.13 &0.9813 &0.0137 &0.0779\\
\checkmark &37.31 &0.9848 &0.0102 &0.0504\\
\bottomrule
\end{tabular}
\caption{Quantitative analysis of the use of additional masks on the HQVI dataset with a 240p input resolution.}
\label{table4}
\end{table}

\begin{table}[t!]
\centering
\small
\begin{tabular}{p{1.8cm}P{1.8cm}P{1.5cm}}
\toprule 
Method &Resolution &Avg. Rank\\
\midrule
FuseFormer &240$\times$432 &2.52\\
ProPainter &480$\times$864 &1.90\\
\rowcolor{Gray} RGVI &480$\times$864 &\textbf{1.59}\\
\hline
\end{tabular}
\caption{User study on the DAVIS 2016 dataset.}
\label{table5}
\end{table}

\subsection{Analysis}
\label{analysis}

\textbf{Pixel propagation.} 
In Table~\ref{table3}, we compare different methods for internal pixel propagation (Int.) and reference propagation (Ref.). One-shot pulling (One.) with bi-directional pixel collection consistently outperforms recurrent warping (Rec.) with sequential pixel distribution~\cite{DFVI, ProPainter} across all evaluation metrics, particularly in perceptual quality measures. This improvement is attributed to one-shot pulling's ability to avoid texture blurring, achieving significantly higher-fidelity content propagation. For recurrent warping, the initial frame is used as the key frame since connection counts cannot be calculated.

\begin{figure}[t]
\centering
\includegraphics[width=1.0\linewidth]{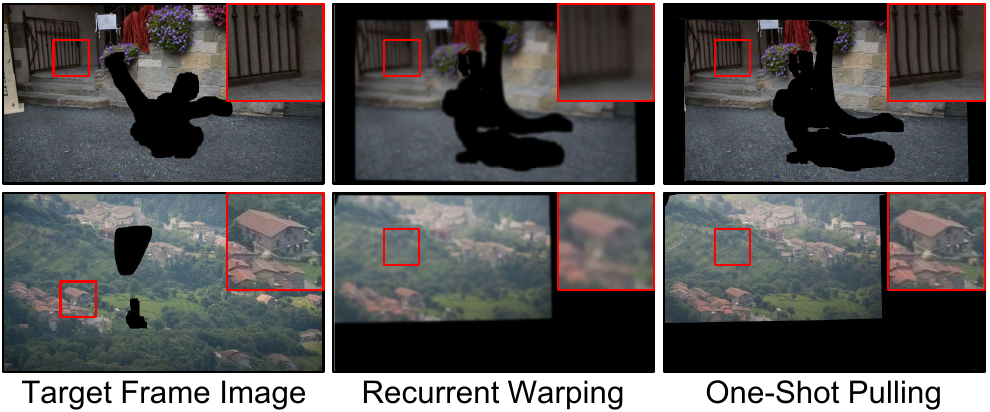}
\caption{Visual comparison between recurrent pixel warping and one-shot pixel pulling, where the source frame image is propagated to the target frame.}
\label{figure6}
\end{figure}

We also present a controlled experiment in Figure~\ref{figure6} to highlight the qualitative effectiveness of our propagation method. Using completed optical flows, we propagate the source frame image to the target frame. The figure illustrates that recurrent pixel warping loses the details of the original content due to repeated pixel sampling. In contrast, our one-shot pixel pulling protocol preserves fine details by warping the original pixels only once.

\begin{figure}[t]
\centering
\includegraphics[width=1.0\linewidth]{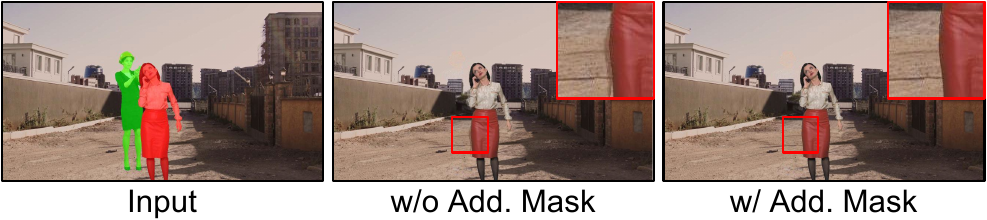}
\caption{Qualitative analysis of the use of additional masks. Negative masks are highlighted in green, while positive masks are highlighted in red.}
\label{figure7}
\end{figure}

\vspace{1mm}
\noindent\textbf{Reference generation.} 
To complete the remaining missing areas after internal pixel propagation, we utilize a large generative model to obtain high-quality reference contents. During this process, we control the generation of new content by providing text prompts. Figure~\ref{figure3} visualizes various texture replacement examples of RGVI, demonstrating that the generation process can be easily controlled by users, highlighting its versatility and applicability.

\vspace{1mm}
\noindent\textbf{Additional masking.} 
The flow-based VI approaches heavily rely on completed optical flows, making them vulnerable in scenarios where the target object is occluded by other distracting objects. To address these limitations, we propose an effective strategy to handle occlusions. By incorporating an additional mask that identifies the occluding object, we mitigate bleeding artifacts, as illustrated in Figure~\ref{figure7}. The quantitative impact of this approach is demonstrated through comparisons in Table~\ref{table4}. Note that evaluations are conducted specifically on videos containing occluding objects.

\vspace{1mm}
\noindent\textbf{User study.} 
Although we perform quantitative evaluations, human evaluation is crucial for comparing various VI methods. Therefore, we conduct a user study on the DAVIS 2016 dataset, using 29 videos sampled from prior work~\cite{dilate}, in Table~\ref{table5}. Based on the average ranking of three methods (FuseFormer, ProPainter, and RGVI) by 10 participants, RGVI demonstrates the highest visual quality by a significant margin.

\vspace{1mm}
\noindent\textbf{Limitations.} 
Similar to other flow-based VI methods, RGVI relies on optical flows. Therefore, inaccuracies in flow completion can result in noticeable structural misalignment in its results. Additionally, the generated reference content may sometimes appear unnatural. Addressing these issues could involve enhancements through the use of additional positive masks and the integration of more robust generative models.

\section{Conclusion}
We present a robust VI system based on a decoupled framework. By enhancing the pixel propagation protocol and integrating powerful generation capabilities, we achieve consistent production of high-fidelity content. The effectiveness of our approach is demonstrated through quantitative and qualitative evaluations. Furthermore, we introduce a novel benchmark dataset tailored for evaluating VI performance in real-world applications. We anticipate that our solution will serve as a potent tool for video editing tasks.

\section{Acknowledgement}
This work was supported by Korea Planning \& Evaluation Institute of Industrial Technology (KEIT) grant funded by the Korea government MOTIE (No. RS-2024-00442120, 50\%) and the National Research Foundation of Korea (NRF) grant funded by the Korea government MSIT (No. RS-2024-00340745, 50\%).

\bibliography{aaai25}
\end{document}